\documentclass[10pt,twocolumn]{article}

\usepackage[margin=0.75in]{geometry}
\usepackage{amsmath, amssymb}
\usepackage{graphicx}
\usepackage{booktabs}
\usepackage{cite}
\usepackage{hyperref}
\usepackage{caption}
\usepackage{cite}
\usepackage{algorithm}
\usepackage{algpseudocode}
\usepackage{multirow}

\title{Kinematic Optimization of Phalanx Length Ratios in Robotic Hands Using Potential Dexterity}

\author{
    HyoJae Kang$^{1}$, Joonho Lee$^{1}$, Jeongdo Ahn$^{1}$, and Dong Il Park$^{1,*}$ \\
    $^{1}$Advanced Robotics Research Center, Korea Institute of Machinery \& Materials (KIMM) \\
    $^{*}$Corresponding author
}

\begin{document}
\date{}
\maketitle

\thispagestyle{plain}
\footnotetext{This manuscript has been submitted for possible publication.}

\begin{abstract}
In the design stage of robotic hands, it is not straightforward to quantitatively evaluate the effect of phalanx length ratios on dexterity without defining specific objects or manipulation tasks. Therefore, this study presents a framework for optimizing the phalanx length ratios of a five-finger robotic hand based on potential dexterity within a kinematic structure. The proposed method employs global manipulability, workspace volume, overlap workspace volume, and fingertip sensitivity as evaluation metrics, and identifies optimal design configurations using a weighted objective function under given constraints. The reachable workspace is discretized using a voxel-based representation, and joint motions are discretized at uniform intervals for evaluation. The optimization is performed over design sets for both the thumb and the other fingers, and design combinations that do not generate overlap workspace are excluded. The results show that each phalanx does not contribute equally to the overall dexterity, and the factors influencing each phalanx are identified. In addition, it is observed that the selection of weighting coefficients does not necessarily lead to the direct maximization of individual performance metrics, due to the non-uniform distribution of evaluation measures within the design space. The proposed framework provides a systematic approach to analyze the trade-offs among reachability, dexterity, and controllability, and can serve as a practical guideline for the kinematic design of multi-fingered robotic hands.
\end{abstract}

\section{Introduction}
The human hand enables a wide range of interactions with the environment through its complex kinematic structure and functional capabilities\cite{c1,c2}. It is regarded as one of the most sophisticated biological systems for performing both power and precision tasks\cite{c3}. Inspired by these characteristics, robotic systems have developed end-effectors that serve similar roles, allowing interaction with objects in diverse environments\cite{c4}. These robotic hands have been widely utilized in applications involving repetitive, labor-intensive, and hazardous tasks\cite{c5,c6,c7}.

To achieve such functionalities, various types of robotic end-effectors have been developed, ranging from low-degree-of-freedom two-finger grippers to multi-finger grippers and anthropomorphic robotic hands. Recently, there has been an increasing demand for robotic systems capable of operating in human-centered environments, where collaboration with humans and handling of objects with diverse geometries are required\cite{c8}. In such scenarios, higher dexterity becomes a key requirement.

Dexterity is widely used to describe the skillfulness of robotic hands. Previous studies\cite{c9} categorize dexterity into three main types: potential dexterity, grasp dexterity, and manipulability dexterity. Potential dexterity refers to the set of reachable configurations without considering object interaction, and is typically characterized by kinematic redundancy and thumb opposability. It reflects the diversity of achievable postures and represents the inherent capability of the hand prior to interaction\cite{c9}. Kinematic redundancy indicates the number of possible joint configurations that can achieve a given pose, while thumb opposability can be evaluated through overlap regions between the thumb and other fingers\cite{c10,c11} or through metrics such as the Kapandji test\cite{c12}.

Grasp dexterity evaluates the set of stable static configurations when grasping objects, typically based on grasp taxonomies\cite{c9,c14,c15,c16}. These approaches focus on the number and diversity of feasible grasp types, and may also consider the ability to switch between different grasping modes\cite{c9}. Manipulability dexterity refers to the ability to reposition and reorient a grasped object within the workspace, reflecting the capability for in-hand manipulation\cite{c1}.

In the design stage of robotic hands, general-purpose functionality is often desired unless a specific task is predefined. However, generality makes it difficult to define representative objects or manipulation tasks for evaluation. Some approaches attempt to address this issue by selecting object sets or primitive manipulation tasks, but these methods still struggle to comprehensively evaluate the overall capability of the hand.

To address this limitation, this paper focuses on potential dexterity, which does not require predefined objects or tasks, and proposes a method for optimizing the kinematic structure of a five-finger robotic hand. Specifically, the optimization problem is formulated to determine the phalanx length ratios of the thumb and fingers within a predefined kinematic structure. The proposed framework evaluates multiple aspects of dexterity while excluding transmission mechanisms and actuation strategies, and instead focuses on structural characteristics such as degrees of freedom (DoF) and link length distributions.

The main contributions of this paper are summarized as follows:

\begin{itemize}
\item A potential dexterity-based optimization framework for designing five-finger robotic hand kinematic structures without relying on object-specific tasks.
\item A systematic evaluation method that integrates workspace, thumb opposability, manipulability, and sensitivity into a unified design criterion.
\item An analysis of how individual design parameters, particularly phalanx length ratios, influence different aspects of dexterity, providing practical design guidelines.
\end{itemize}

The remainder of this paper is organized as follows. Section 2 reviews existing research on five-finger robotic hands and related evaluation methods. Section 3 introduces the kinematic structure of the robotic hand. Section 4 describes the proposed optimization algorithm. Section 5 presents the optimization process, and Section 6 provides the results and analysis.

\section{Related Works}

Stable grasping of objects typically requires at least three contact points, which corresponds to the use of three or more fingers\cite{c18}. However, for more complex tasks such as in-hand manipulation and dexterous operations, a larger number of fingers and more sophisticated structures are often required. To address these demands, various robotic end-effectors have been developed, including anthropomorphic hands, in-hand manipulation-oriented grippers\cite{c19}, and reconfigurable or multi-modal grippers capable of adapting to different tasks\cite{c20,c21}.

Compared to conventional grippers, dexterous robotic hands provide significant advantages in human-centered environments. In particular, their ability to handle objects designed for human use makes them suitable for collaborative applications. Numerous studies have explored five-finger anthropomorphic robotic hands\cite{c22,c23,c24,c25,c26,c27,c28,c29}, demonstrating a wide range of actuation methods, transmission mechanisms, and DoF configurations.

The performance of these robotic hands has been evaluated using various approaches. Experimental validation based on grasp taxonomies is commonly used to assess the range of feasible grasp types. In addition, in-hand manipulation experiments, motion-based evaluations such as the Kapandji test, and force or manipulability analyses have been employed to verify performance. These approaches provide meaningful insights into the functional capabilities of robotic hands and are widely adopted in existing studies.

However, most of these approaches focus on post-design evaluation, where the performance is assessed after the hand has already been developed. Relatively less attention has been given to the question of how the kinematic structure itself should be designed in the early stages, particularly when no specific task or object is predefined.

From a design perspective, key factors related to potential dexterity, such as workspace, thumb opposability, and manipulability, are strongly influenced by the kinematic structure of the hand, including phalanx length ratios, DoF, and joint motion ranges. Therefore, optimizing these structural parameters in the absence of predefined tasks can provide a more general and robust foundation for designing high-performance robotic hands.

In this context, this paper proposes a method for optimizing the potential dexterity of a robotic hand composed of four DoF fingers and a five DoF thumb. The proposed approach not only evaluates the overall dexterity but also analyzes how individual design parameters affect different performance metrics, providing insights into the relationship between kinematic structure and dexterity.

\section{Kinematic Structure of Dexterous Hand}

In this section, the kinematic structure of the five-finger robotic hand to be evaluated in this study is introduced. The proposed hand consists of four 4-DoF fingers and one 5-DoF thumb. Each 4-DoF finger is composed of three DoFs for flexion/extension (F/E) motion and one DoF for adduction/abduction (A/A) motion, while the 5-DoF thumb is composed of three DoFs for F/E motion, one DoF for A/A motion, and one DoF for opposition/reposition (O/R) motion.

\begin{figure}[!t]
    \centering
    \includegraphics[width=0.99\columnwidth]{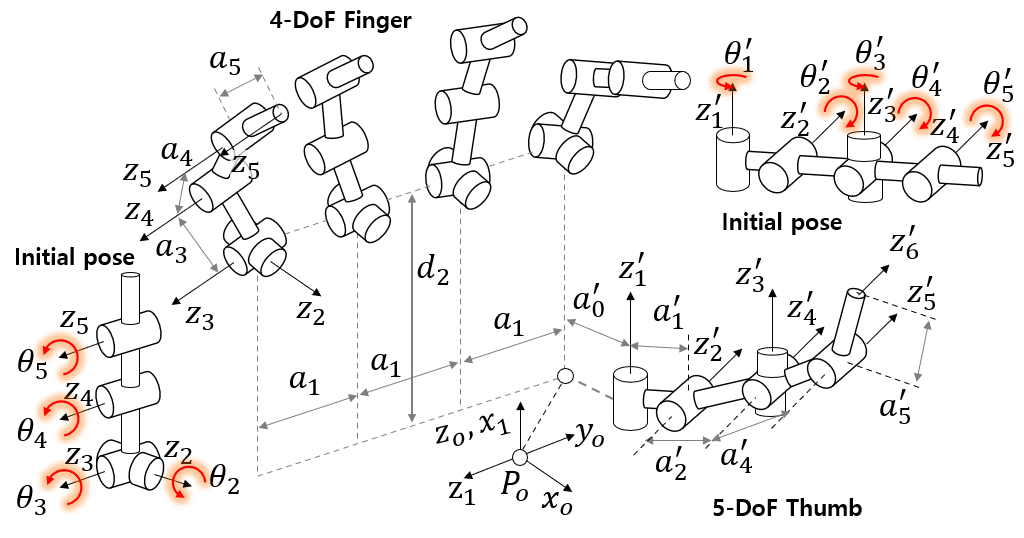}
    \caption{Kinematic structure and initial configuration of the five-finger robotic hand considered in this study}
    \label{Fig1}
\end{figure}

Figure ~\ref{Fig1} shows the kinematic structure of the five-finger dexterous hand. Although various hand configurations have been studied in previous works, the objective of this study is not to identify the best overall configuration, but to present a method for optimizing the phalanx lengths in a given configuration. Therefore, the structure shown in Figure ~\ref{Fig1}, which implements multiple motions except for palm motion, was adopted. Accordingly, depending on the designer’s research purpose, it is necessary to adjust the kinematic parameters, offsets, and initial installation angles.

Figure ~\ref{Fig1} illustrates the structural arrangement of each finger. The initial postures of the thumb and the other fingers, as well as the coordinate frames and angular variables of each joint ($\theta_i, \theta_i'$), are also indicated.

\begin{figure}[!t]
    \centering
    \includegraphics[width=0.95\columnwidth]{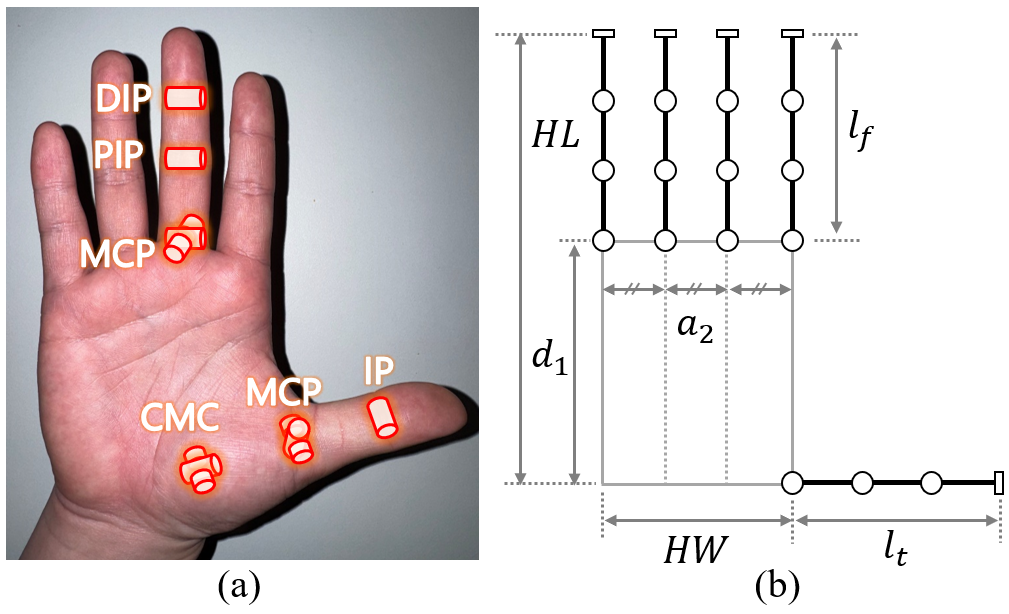}
    \caption{Kinematic structure of the hand (a) Joint configuration of a human hand, (b) Key parameters of the simplified structure}
    \label{Fig2}
\end{figure}

Figure~\ref{Fig2}(a) illustrates a human hand, with the joints of each finger and the thumb explicitly indicated. For the fingers, the joints are defined—starting from the fingertip—as the distal interphalangeal (DIP), proximal interphalangeal (PIP), and metacarpophalangeal (MCP) joints. For the thumb, the corresponding joints consist of the interphalangeal (IP), MCP, and carpometacarpal (CMC) joints.

Figure~\ref{Fig2}(b) presents the key geometric parameters obtained after structural simplification of the hand model. In this study, for the purpose of structural abstraction, the index to little fingers are assumed to share identical kinematic parameters. Furthermore, the thumb is modeled to be longer than the other fingers, and a uniform spacing between adjacent fingers is maintained.

In the proposed kinematic structure, the overall hand length is normalized to 1. Based on this, the length of each phalanx is expressed as a ratio.  Let the thumb length be denoted by $l_t$ and the length of the other fingers by $l_f$. The thumb was set to be at least 10\% longer than the index finger, and the dimensions were discretized in units of 0.01 so that the total length could be evenly distributed into three phalanges. Let the distance from the index finger to the little finger be defined as hand width (HW). Then, by taking the average ratio of hand width to hand length (HL) in study \cite{c31}, the following relationship is obtained:

\begin{subequations}
\begin{align}
    \mathrm{HW} = 3a_2,\\
    \mathrm{HL} = d_1 + l_f,\\
    \mathrm{HW} = 0.54 \mathrm{HL}.
\end{align}
\end{subequations}

In this study, the design variables of primary interest correspond to the phalanges in which F/E motion occurs. As shown in Figure ~\ref{Fig1}, these correspond to $a_3, a_4, a_5, a_2', a_4'$, and $a_5'$, whereas other offset dimensions such as $a_0'$ and $a_1'$ were set arbitrarily. The resulting kinematic parameters are summarized in Table ~\ref{Tab2}.

\begin{table}[h] 
\centering
\caption{Kinematic parameter lengths of the five-finger robotic hand, with the hand length normalized to 1.}
\label{Tab2}
\begin{tabular}{c c c c c c c c}
\midrule
HL & HW & $l_t$ & $l_f$ & $a_1$ & $a_0'$ & $a_1'$ & $d_1$ \\
\midrule
1 & 0.54 & 0.51 & 0.45 & 0.18 & 0.10 & 0.10 & 0.55 \\
\midrule
\end{tabular}
\end{table}

Based on the above description, the coordinates of each fingertip can be represented with respect to the base frame $\{P_o, x_o, y_o, z_o\}$. The fingertip position was computed using homogeneous transformation matrices. The transformation from link $i$ to $i+1$ is expressed as:

\begin{equation}
    T_i^{i+1} = 
    \begin{bmatrix}
 R_i^{i+1} & p_i^{i+1} \\
 0 & 1 \\
\end{bmatrix}.
\end{equation}

where $R_i^{i+1}$ is the rotation matrix and $p_i^{i+1}$ is the displacement vector between links. The overall transformation matrix is obtained by multiplying the transformations of each link:

\begin{equation}
    T_i^j = T_i^{i+1} T_{i+1}^{i+2} \cdots T_{j-2}^{j-1} T_{j-1}^j.
\end{equation}

Each transformation matrix can be derived using the DH parameter tables in Tables ~\ref{Tab3} and ~\ref{Tab4}. Table ~\ref{Tab3} presents the five DoF thumb, and Table ~\ref{Tab4} presents the four DoF middle finger. For the fingers other than the middle finger, the structure can be represented by modifying the offsets. Using the DH parameter convention, the joints of the fingers are represented as being interconnected through a kinematic chain of links \cite{c32}. By multiplying each transformation matrix, the transformation from frame 0 to the fingertip can be obtained. Using this, the fingertip position $p^e$ with respect to the base frame $p_o x_o y_o z_o$ is computed as:

\begin{equation}
    p^e = T_0^e p^o.
\end{equation}

\begin{table}[h] 
\centering
\caption{DH parameters of the five DoF thumb}
\label{Tab3}
\begin{tabular}{c c c c c}
\midrule
$i$ & $\alpha_{i-1}$ & $a_{i-1}$ & $d_i$ & $\theta_i$ \\
\midrule
1 & $0^\circ$    & $a_0'$ & 0 & $\theta_1'$ \\
2 & $-90^\circ$  & $a_1'$ & 0 & $\theta_2'$ \\
3 & $90^\circ$   & $a_2'$ & 0 & $\theta_3'$ \\
4 & $-90^\circ$  & 0      & 0 & $\theta_4'$ \\
5 & $0^\circ$    & $a_4'$ & 0 & $\theta_5'$ \\
6 & $0^\circ$    & $a_5'$ & 0 & 0 \\
\midrule
\end{tabular}
\end{table}

\begin{table}[h] 
\centering
\caption{DH parameters of the four DoF middle finger}
\label{Tab4}
\begin{tabular}{c c c c c}
\midrule
$i$ & $\alpha_{i-1}$ & $a_{i-1}$ & $d_i$ & $\theta_i$ \\
\midrule
1 & $90^\circ$   & 0     & 0   & $90^\circ$ \\
2 & $90^\circ$   & $a_1$ & $d_2$ & $\theta_2$ \\
3 & $-90^\circ$  & 0     & 0   & $\theta_3$ \\
4 & $0^\circ$    & $a_3$ & 0   & $\theta_4$ \\
5 & $0^\circ$    & $a_4$ & 0   & $\theta_5$ \\
6 & $0^\circ$    & $a_5$ & 0   & 0 \\
\midrule
\end{tabular}
\end{table}

In addition, the joint angle ranges of the thumb and the other fingers were set as shown in Tables ~\ref{Tab5} and ~\ref{Tab6}, respectively.

\begin{table}[h] 
\centering
\caption{Range of motion settings for each actuated joint of the thumb}
\label{Tab5}
\begin{tabular}{c c c c}
\midrule
$\theta_1'$ & $\theta_2'$ & $\theta_3'$ & $\theta_4'$ and $\theta_5'$ \\
\midrule
$-30^\circ$ to $90^\circ$ &
$-90^\circ$ to $30^\circ$ &
$-60^\circ$ to $60^\circ$ &
$-90^\circ$ to $0^\circ$ \\
\midrule
\end{tabular}
\end{table}

\begin{table}[h] 
\centering
\caption{Range of motion settings for each actuated joint of the fingers}
\label{Tab6}
\begin{tabular}{c c c}
\midrule
$\theta_2$ & $\theta_3$ & $\theta_4$ and $\theta_5$ \\
\midrule
$-30^\circ$ to $30^\circ$ &
$-90^\circ$ to $30^\circ$ &
$-90^\circ$ to $0^\circ$ \\
\midrule
\end{tabular}
\end{table}

Based on these settings, the reachable regions of the thumb and the other fingers can be identified and analyzed according to the joint angle ranges. In addition, the reachable workspace and kinematic characteristics vary depending on the length ratio of each phalanx.

\section{Potential Dexterity Evaluation Factors}

In this study, a sampling-based analysis method on a joint grid was employed to quantitatively evaluate the potential dexterity based on the kinematic structure of the finger. The analysis process consists of five main steps: (1) workspace sampling based on forward kinematics, (2) manipulability computation, (3) voxel-based workspace volume estimation, (4) workspace overlap computation between fingers, and (5) sensitivity evaluation of the finger.

\subsection{Workspace Sampling}

In this study, the workspace of the fingertip was evaluated by sampling on a discretized joint grid. The fingertip position is computed using forward kinematics derived from the kinematic structure described in Section 4. Each joint variable is sampled uniformly within its predefined range:

\begin{equation}
    q_i \in {q_{i,1}, q_{i,2}, \cdots , q_{i,n_i}}, q_{i,k}=q_{i,\min}+(k-1)\Delta q_i.
\end{equation}

where $n_i$ is the number of discretization steps and $\Delta q_i$ is the step size. The overall joint configuration set $\mathcal{Q}$ is constructed as a Cartesian product, and a workspace point set is generated for all sampled configurations:

\begin{equation}
    P={p_1,p_2,\cdots, p_{N_q}}.
\end{equation}

This process is independently applied to the thumb and each of the other four fingers, and Algorithm A~\ref{algo1} shows the workspace sampling process.

\begin{algorithm} 
\caption{A1 Workspace Sampling using Joint Grid}
\label{algo1}
\begin{algorithmic}[1]
\Require Design parameters $d$, joint grid $\mathcal{Q}$
\Ensure Workspace point set $P$

\State Initialize $P \leftarrow \emptyset$

\For{each $q \in \mathcal{Q}$}
    \State Compute fingertip position $p^e = f(q; d)$ using forward kinematics
    \State $P \leftarrow P \cup \{p^e\}$
\EndFor

\State \Return $P$
\end{algorithmic}
\end{algorithm}

\subsection{Manipulability}

The kinematic manipulability at each joint configuration was computed using manipulability \cite{c43}. The Jacobian matrix $J(q)$ relates the fingertip velocity and joint velocity as:

\begin{equation}
    \Dot{p}^e = J(q)\Dot{q}.
\end{equation}

The manipulability is defined as:

\begin{equation}
    w(q) = \sqrt{\det (JJ^T)}.
\end{equation}

In this study, the manipulability was computed using the analytical Jacobian, and the global manipulability was defined as the average value over all sampled configurations on the joint grid:

\begin{equation}
    w_g = \frac{1}{N_q}\sum_{q\in \mathcal{Q}} w(q).
\end{equation}

Algorithm A~\ref{algo2} shows the manipulability evaluation process described above.

\begin{algorithm} 
\caption{A2 Global Manipulability Evaluation}
\label{algo2}
\begin{algorithmic}[1]
\Require Design parameters $d$, joint grid $\mathcal{Q}$
\Ensure Global manipulability $w_g$

\State $s \leftarrow 0$, $N \leftarrow 0$

\For{each $q \in \mathcal{Q}$}
    \State Compute Jacobian $J(q)$
    \State Compute $w(q) = \sqrt{\det(JJ^T)}$
    \State $s \leftarrow s + w(q)$
    \State $N \leftarrow N + 1$
\EndFor

\State $w_g \leftarrow \frac{s}{N}$
\State \Return $w_g$
\end{algorithmic}
\end{algorithm}

When the manipulability evaluation result is denoted as $z_1$, the index finger to the little finger share the same design dimension. Therefore, the global manipulability of the thumb $W_{g,t}$ and the global manipulability of the index finger $W_{g,i}$ are equally weighted as:

\begin{equation}
    z_1 = 0.5 W_{g,t} + 0.5 W_{g,i}.
\end{equation}

\subsection{Workspace volume}

Workspace volume was estimated using a voxel-based discretization approach, where occupied voxels were identified from sampled workspace points and summed to obtain the total volume\cite{c41}. A study evaluating the reachable workspace of manipulators using voxel-based discretization of joint motions has been presented \cite{c42}. The workspace volume was computed using a voxel-based space partitioning method.

First, the workspace point set was divided into a voxel grid, and for each workspace point $p = (x,y,z)$, the voxel index is defined as:

\begin{equation}
    v_x = \lfloor \frac{x}{\Delta} \rfloor, v_y = \lfloor \frac{y}{\Delta} \rfloor, v_z = \lfloor \frac{z}{\Delta} \rfloor.
\end{equation}

Each point is mapped to a voxel index $v = (v_x, v_y, v_z)$, and points belonging to the same voxel are treated as a single spatial cell. Let $V_e$ denote the set of occupied voxels, then the workspace volume is computed as:

\begin{equation}
    V_w = |V_e|\Delta^3.
\end{equation}

Algorithm A~\ref{algo3} shows the workspace volume estimation process. 

\begin{algorithm} 
\caption{A3 Workspace Volume using Voxelization}
\label{algo3}
\begin{algorithmic}[1]
\Require Workspace point set $P$, voxel size $\Delta$
\Ensure Workspace volume $V_w$, occupied voxel set $V_e$

\State $V_e \leftarrow \emptyset$

\For{each $p = (x,y,z) \in P$}
    \State $v_x \leftarrow \lfloor x/\Delta \rfloor$
    \State $v_y \leftarrow \lfloor y/\Delta \rfloor$
    \State $v_z \leftarrow \lfloor z/\Delta \rfloor$
    \State $V_e \leftarrow V_e \cup \{(v_x, v_y, v_z)\}$
\EndFor

\State $V_w \leftarrow |V_e| \Delta^3$
\State \Return $V_w, V_e$
\end{algorithmic}
\end{algorithm}

The workspace volume term $z_{2w}$ is defined as the weighted sum of the workspace volumes of the thumb and the index finger, with equal weighting applied to both, as follows:

\begin{equation}
    z_{2w} = 0.5 V_{w,t} + 0.5 V_{w,i}.
\end{equation}

\subsection{Overlap workspace volume}

To analyze the region that can be simultaneously reached by the thumb and other fingers, the workspace overlap was computed. Let $V_t$ be the voxel set of the thumb workspace, and $V_f$ be the voxel set of the other fingers $(f=i,m,r,l)$. The overlap voxel set $V_{ovr,f}$ and overlap volume $V_{ovr,f}^{vol}$ are defined as:

\begin{subequations}
\begin{align}
    V_{ovr,f} = V_t \cap V_f,\\
    V_{ovr,f}^{vol} = |V_{ovr,f}| \Delta^3.
\end{align}
\end{subequations}

This value represents the volume of space that can be simultaneously reached by two fingers. Algorithm A~\ref{algo4} shows the overlap computation process.

\begin{algorithm}
\caption{A4 Workspace Overlap Computation}
\label{algo4}
\begin{algorithmic}[1]
\Require Thumb voxel set $V_t$, finger voxel set $V_f$
\Ensure Overlap volume $V_{ovr,f}^{vol}$ and total overlap $V_{ovr}^{vol}$

\State $V_{ovr}^{vol} \leftarrow 0$

\For{each finger $f \in \{i,m,r,l\}$}
    \State $V_{ovr,f} \leftarrow V_t \cap V_f$
    \State $V_{ovr,f}^{vol} \leftarrow |V_{ovr,f}| \Delta^3$
    \State $V_{ovr}^{vol} \leftarrow V_{ovr}^{vol} + V_{ovr,f}^{vol}$
\EndFor

\State \Return $V_{ovr,f}^{vol}, V_{ovr}^{vol}$
\end{algorithmic}
\end{algorithm}

The overlap workspace volume term $z_{2o}$ is defined as the sum of the overlap volumes between the thumb and each of the other fingers, as follows:

\begin{equation}
    z_{2o} = 0.25 \sum V_{ovr,f}^{vol}.
\end{equation}

Since both maximizing workspace volume and improving thumb opposability are related to workspace properties, the final workspace-related term is defined as:

\begin{equation}
    z_2 = 0.5 z_{2w} + 0.5 z_{2o}.
\end{equation}

\subsection{Sensitivity}

Sensitivity is defined to quantify how strongly the fingertip position responds to actuation at the distal F/E joint. To isolate this effect, only F/E motions were considered, while all non-flexion joints were fixed. For each sampled configuration, the analytical Jacobian was computed, and the norm of the column corresponding to the distal joint was used as the instantaneous sensitivity measure:

\begin{equation}
    s_d(q) = \|J_{(:,k)(q)}\|.
\end{equation}

where $k$ denotes the distal joint index. The sensitivity was obtained by averaging this value over all sampled F/E configurations $\mathcal{Q}_{fe}$:

\begin{equation}
    \bar{s}_d = \frac{1}{|\mathcal{Q}_{fe}|}\sum_{q\in \mathcal{Q}_{fe}} ||J_{(:,k)}(q)||.
\end{equation}

This procedure corresponds to Algorithm A~\ref{algo5}, where the sensitivity is accumulated over the sampled configuration set and normalized by the total number of samples.

\begin{algorithm} 
\caption{A5 Sensitivity Evaluation}
\label{algo5}
\begin{algorithmic}[1]
\Require Design parameters $d$, configuration set $\mathcal{Q}_{\mathrm{FE}}$
\Ensure Average distal sensitivity $\bar{s}_d$

\State $S \leftarrow 0$, $N \leftarrow 0$

\For{each $q \in \mathcal{Q}_{\mathrm{FE}}$}
    \State Fix all non-flexion joints
    \State Compute Jacobian $J(q,d)$
    \State $s_d(q) \leftarrow \|J_{(:,k)}(q)\|$
    \State $S \leftarrow S + s_d(q)$
    \State $N \leftarrow N + 1$
\EndFor

\State $\bar{s}_d \leftarrow S/N$
\State \Return $\bar{s}_d$
\end{algorithmic}
\end{algorithm}

The sensitivity evaluation term $z_3$ is defined as follows:

\begin{equation}
    z_3 = \bar{s}_d.
\end{equation}

\section{Phalanx Length Ratio Optimization}

The goal of the optimization problem is to determine the dimensions of the hand elements while simultaneously optimizing multiple objective functions under given constraints. The design variables are defined as:

\begin{equation}
    x = (a_3, a_4, a_5, a_2', a_4', a_5').
\end{equation}

where $a_3, a_4, a_5, a_2', a_4', a_5'$ are the hand design parameters used in the kinematic modeling. Based on these design variables, the global manipulability, workspace volume, overlap workspace volume, and sensitivity are obtained. Therefore, the objective function $f(x)$ is defined by assigning weighting coefficients $c_1, c_2, c_3$ to each evaluation term as follows:

\begin{equation}
    f(x) = \frac{c_1z_1}{z_m} + \frac{c_2z_2}{z_w} - \frac{c_3z_3}{z_s}.
\end{equation}

The objective function is formulated such that larger values of workspace volume and manipulability increase the objective value, while smaller values of sensitivity increase the objective value. Here, $z_m$ denotes the maximum manipulability achievable within the given finger length constraints, $z_w$ denotes the sum of the maximum workspace volume $z_{2w}$ and the maximum overlap workspace volume $z_{2o}$, and $z_s$ denotes the maximum fingertip sensitivity. These values are used to normalize each evaluation term. 

The goal is to derive the optimal length ratios of the design variables that maximize the objective function under different weighting coefficients. Since the weighting coefficients represent relative importance, they satisfy:

\begin{equation}
    c_1 + c_2 + c_3 = 1
\end{equation}

In this study, the weighting coefficients are adjusted to investigate how emphasizing specific evaluation factors affects the resulting optimal design.

The constraints are defined such that the phalanx lengths decrease monotonically toward the fingertip. In addition, the total length of the index finger is set to 0.45, and the total length of the thumb is set to 0.51. Each phalanx is constrained to have a minimum length of 0.10 to prevent degenerate configurations.

\begin{subequations}
\begin{align}
    a_3 + a_4 + a_5 = 0.45, a_3 \geq a_4 \geq a_5 \geq 0.10,\\
    a_2' + a_4' + a_5' = 0.51, a_2' \geq a_4' \geq a_5' \geq 0.10.
\end{align}
\end{subequations}
These constraints can be rewritten as:

\begin{subequations}
\begin{align}
    a_3 = 0.45 - a_4 - a_5,\\
    a_2' = 0.51 - a_4' - a_5'.
\end{align}
\end{subequations}

\begin{algorithm}[!t]
\caption{A6 Searching for Optimal Design Pair}
\label{algo7}
\begin{algorithmic}[1]
\Require Thumb design set $D_t$, Finger design set $D_f$, voxel size $\Delta$, weight coefficients $c_1, c_2, c_3$, normalization values $z_m, z_w, z_s$, Thumb joint grid $\mathcal{Q}_t$, Finger joint grid $\mathcal{Q}_f$
\Ensure Optimal thumb design $d_t^*$, Optimal finger design $d_f^*$, Maximum objective value $f^*$

\State Generate feasible design set of thumb $D_t$ and finger $D_f$

\For{each $d_t \in D_t$}
    \State Generate workspace point set using A~\ref{algo1}
    \State Compute global manipulability using A~\ref{algo2}
    \State Compute workspace volume and voxel set using A~\ref{algo3}
    \State Compute distal sensitivity using A~\ref{algo5}
    \State Store $(d_t, V_t, W_t, V_t^{vox}, S_t)$
\EndFor

\For{each $d_f \in D_f$}
    \State Generate workspace point set using A~\ref{algo1}
    \State Compute global manipulability using A~\ref{algo2}
    \State Compute workspace volume and voxel set using A~\ref{algo3}
    \State Compute distal sensitivity using A~\ref{algo5}
    \State Store $(d_f, V_f, W_f, V_f^{vox}, S_f)$
\EndFor

\State Initialize $f^* \leftarrow -\infty$

\For{each stored thumb result $r_t$}
    \For{each stored finger result $r_f$}
        \State Compute overlap volume using A~\ref{algo4}
        \If{any overlap volume is zero}
            \State \textbf{continue}
        \EndIf

        \State Compute normalized objective:
        \[
        f(x) = \frac{c_1 z_1}{z_m} + \frac{c_2 z_2}{z_w} - \frac{c_3 z_3}{z_s}
        \]

        \If{$f(x) > f^*$}
            \State $f^* \leftarrow f(x)$
            \State $d_t^* \leftarrow d_t$, $d_f^* \leftarrow d_f$
        \EndIf
    \EndFor
\EndFor

\State \Return $d_t^*, d_f^*, f^*$
\end{algorithmic}
\end{algorithm}

Next, the optimization procedure is described. Let $D$ denote the set of feasible designs satisfying the finger length constraints. For each design candidate $d\in D$, the workspace point set and global manipulability are computed using Algorithm A~\ref{algo1} and A~\ref{algo2}. The voxelized workspace volume is then computed from the generated workspace point set using Algorithm A~\ref{algo3}.

The same procedure is applied to the thumb and the other four fingers. Using the voxel sets obtained from each design, the overlap workspace volumes between the thumb and each of the four fingers are computed using Algorithm A~\ref{algo4}. Design combinations that produce zero overlap workspace volume are excluded from further consideration.

Based on this process, the optimization procedure for maximizing the objective function is described in Algorithm A~\ref{algo7}. For each design configuration, the workspace volume, global manipulability, and voxel sets are stored. This process is applied to both the thumb and the four fingers. Then, for each combination of thumb and finger designs, the overlap workspace volume is computed. The evaluation terms are normalized using their corresponding maximum values, and the objective function is evaluated. Finally, the optimal design combination that maximized the objective function under the given weighting coefficients is selected.

The maximum values $z_m, z_w, z_s$ used for normalization can be obtained by setting the weighting coefficients in Algorithm A~\ref{algo7} to consider only a single evaluation term at a time. In this process, the minimum length constraints are applied while removing the phalanx length ordering constraints. Through this approach, the normalization values are computed globally, while the actual optimization process is performed under the full set of constraints.

\section{Results and Analysis}

In this section, the pre-defined settings used for the optimization are described, and the optimization results obtained under different weighting coefficients are presented. In addition, the influence of each evaluation factor on the finger phalanx length ratios is analyzed based on the results. Furthermore, the relationships between the evaluation factors and the resulting design tendencies are discussed.

\subsection{Voxel and Resolution}

The voxel size $\Delta$ must be determined for workspace and overlap workspace analysis. To determine $\Delta$, an appropriate resolution is required. The resolution refers to how finely the joint motion ranges are discretized. In this study, both the resolution and $\Delta$ were determined by observing the variation in the results after setting all phalanx lengths corresponding to the design variables to equal values. The resolution was selected at the point where the variation became less than 2\%.

When the total hand length is set to 1, the thumb phalanx lengths are assigned as 0.17, 0.17, and 0.17 for the metacarpal, proximal, and distal segments, respectively. The phalanx lengths of the other fingers are assigned as 0.15, 0.15, and 0.15 for the proximal, middle, and distal segments, respectively. The workspace volume was then evaluated while decreasing the joint angle resolution from 30$^\circ$.

At this stage, the voxel size was set to four different values: 0.05, 0.025, 0.01, and 0.005. The significant points were observed at resolutions of 5$^\circ$ and 3$^\circ$ for the thumb with voxel sizes of 0.05 and 0.025, and at resolutions of 3$^\circ$ and 2$^\circ$ for the fingers with the same voxel sizes.

\begin{table}[h] 
\centering
\caption{Workspace volumes of the thumb and fingers at different sampling resolutions}
\label{Tab55}
\begin{tabular}{c|cc|cc}
\midrule
Voxel & \multicolumn{2}{c|}{Thumb} & \multicolumn{2}{c}{Finger} \\
  ($\Delta$)& 5$^\circ$ & 3$^\circ$ & 3$^\circ$ & 2$^\circ$ \\
\midrule
0.05  & 0.35825 & 0.36388 & 0.08575 & 0.08600 \\
0.025 & 0.29347  & 0.30130 & 0.06245 & 0.06295 \\
\midrule
\end{tabular}
\end{table}

Table ~\ref{Tab55} shows the workspace volume according to voxel size and resolution. The results indicate that, for $\Delta=0.05$, the change in volume remains within 2\% for both the thumb and the fingers as the resolution changes. As the voxel size decreases, the occupied volume decreases, whereas larger voxel sizes assign larger volumes even with fewer occupied points. Additionally, higher resolution leads to a larger computed volume due to denser sampling; however, the variation becomes negligible beyond a certain point. Therefore, the voxel size was set to $\Delta = 0.05$, with a resolution of 5$^\circ$ for the thumb and 3$^\circ$ for the other fingers.

\subsection{Optimization Results}

Before performing the optimization, the number of feasible designs is determined by applying the minimum length constraint and the ordering constraint among phalanx lengths. As a result, there are 48 possible thumb designs and 27 possible designs for the other fingers.

The optimization results obtained by evaluating all combinations of these design candidates are presented. First, seven combinations of weighting coefficients were defined, as shown in Table ~\ref{Tab66}. Case 1 corresponds to equal weighting, Case 2, 3 and 4 emphasize a single evaluation factor, and the remaining three cases (Case 5, 6, and 7) emphasize two evaluation factors.

\begin{table}[h] 
\centering
\caption{Classification of cases defined by seven different weighting coefficients}
\label{Tab66}
\begin{tabular}{c|ccc}
\midrule
Case & $c_1$ & $c_2$ & $c_3$ \\
\midrule
1 & 0.33 & 0.33 & 0.33 \\
2 & 0.8  & 0.1  & 0.1  \\
3 & 0.1  & 0.8  & 0.1  \\
4 & 0.1  & 0.1  & 0.8  \\
5 & 0.4  & 0.4  & 0.2  \\
6 & 0.4  & 0.2  & 0.4  \\
7 & 0.2  & 0.4  & 0.4  \\
\midrule
\end{tabular}
\end{table}

\begin{figure}[!t]
    \centering
    \includegraphics[width=0.95\columnwidth]{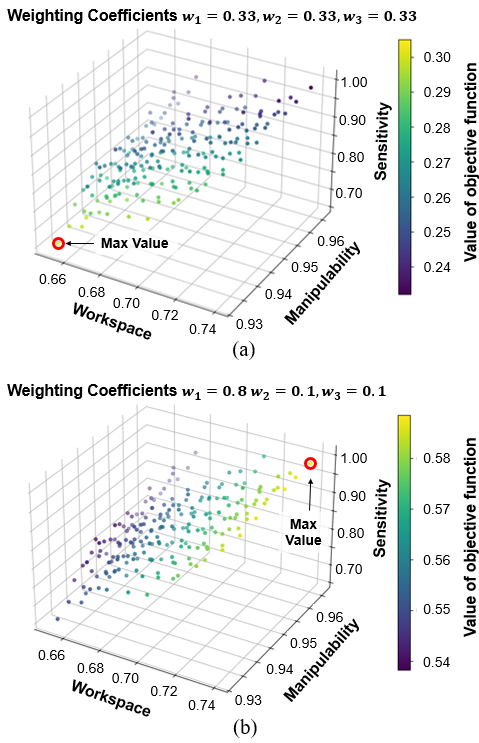}
    \caption{Distribution of workspace volume, manipulability, and sensitivity for each design satisfying the optimization constraints, along with the corresponding objective function values under different weighting coefficients: (a) Case 1 - $w_1=w_2=w_3=0.33$, (b) Case 2 - $w_1=0.8, w_2=w_3=0.1$}
    \label{Fig5}
\end{figure}

In Case 1, where the weighting coefficients are equal, design configurations with zero overlap between the thumb and the other fingers were excluded. The distribution of the objective function is shown in Figure ~\ref{Fig5}(a).

Within the feasible design space, the variation of workspace and manipulability is relatively smaller compared to that of sensitivity. Since workspace and manipulability are proportional to the objective function and sensitivity is inversely proportional, the objective function is observed to be most sensitive to the variation of sensitivity.

Figure ~\ref{Fig5}(b) shows Case 2, where the weighting coefficient for workspace is increased while those for manipulability and sensitivity are reduced. In this case, the variation in workspace dominates the objective function by a factor of approximately eight compared to the other terms, resulting in a design biased toward maximizing workspace volume.

\begin{figure}[!t]
    \centering
    \includegraphics[width=0.95\columnwidth]{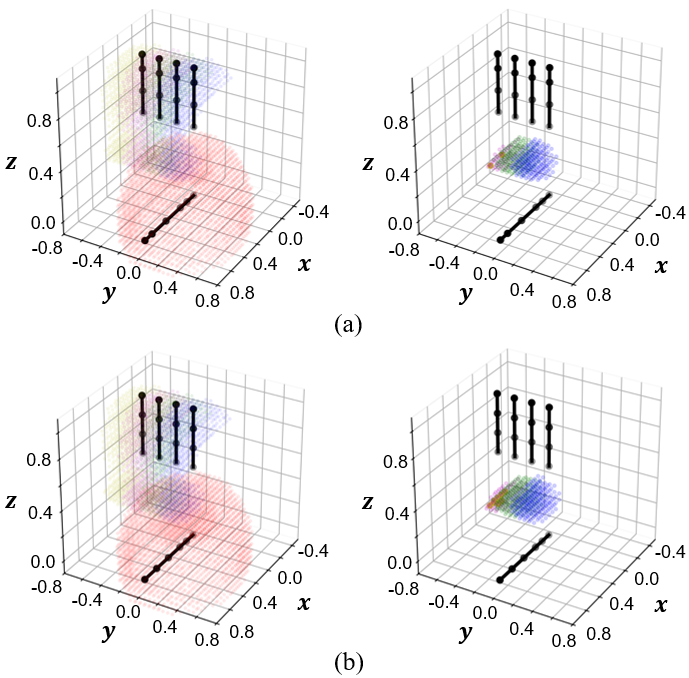}
    \caption{Workspace and overlap workspace corresponding to the designs with the highest objective function values: (a) Case 1, (b) Case 2}
    \label{Fig6}
\end{figure}

Figure ~\ref{Fig6} presents the workspace and overlap workspace for Cases 1 and 2. Although no significant visual difference is observed in Fig. ~\ref{Fig6}(a) and (b), the actual workspace volume increased by 8.27\% for the thumb and 3.21\% for the finger, while the overlap volume increased by 20.96\%.

\begin{figure}[!t]
    \centering
    \includegraphics[width=0.95\columnwidth]{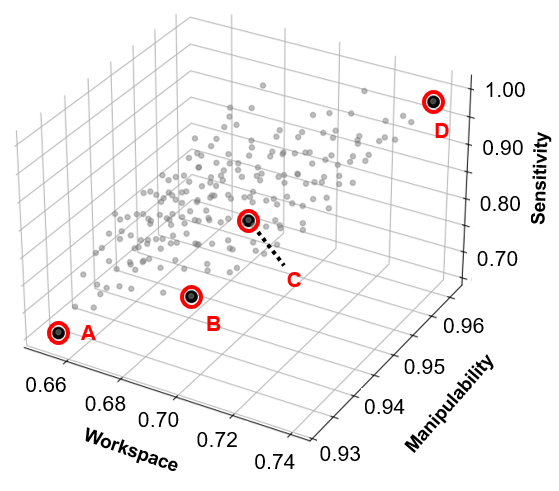}
    \caption{Distribution of objective function values for each design under seven different weighting coefficient configurations, along with four representative optimized results (A, B, C, and D)}
    \label{Fig7}
\end{figure}

The overall results for Cases 1 to 7 are summarized in a single distribution, as shown in Figure ~\ref{Fig7}. Although not all possible weighting combinations were exhaustively explored, the seven cases can be categorized into four distinct groups. In Figure ~\ref{Fig7}, group A corresponds to Cases 1, 4, 6, and 7, group B corresponds to Case 5, group C corresponds to Case 3, and group D corresponds to Case 2.

Cases 1, 4, 6, and 7 exhibit behavior dominated by sensitivity, which has relatively larger variation. Cases with high weighting on workspace, manipulability, or both show distinct results. In particular, when comparing Case 3 (high manipulability weight) with Case 1 (equal weights), the manipulability increased by 0.82\% for the thumb and 3.30\% for the finger. Although Case 2 yields higher manipulability values, this is influenced by the relative variation magnitudes of each evaluation factor within the feasible design space.

\begin{figure}[!t]
    \centering
    \includegraphics[width=0.95\columnwidth]{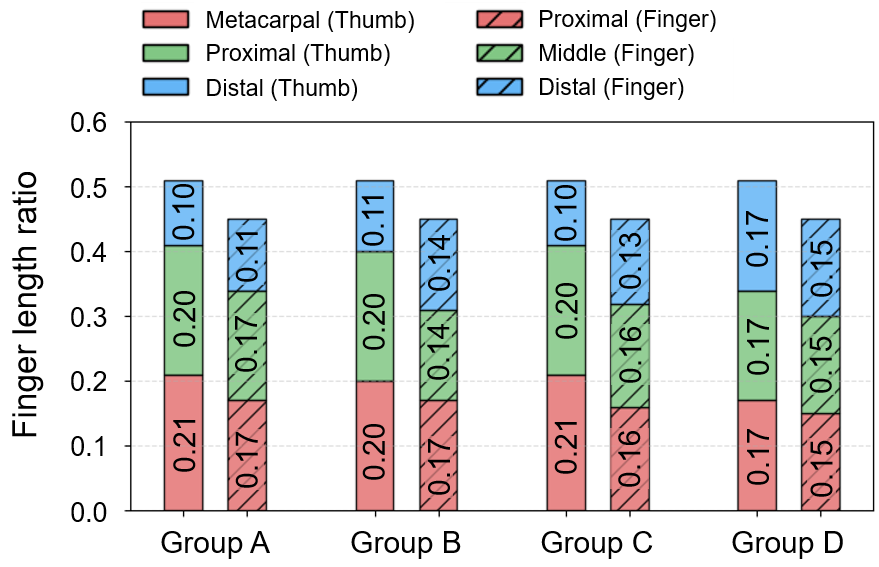}
    \caption{Phalanx length ratios of the thumb and fingers for the four optimized design groups (A, B, C, and D)}
    \label{Fig8}
\end{figure}

Figure ~\ref{Fig8} shows the phalanx length ratios of the thumb and fingers for groups A to D. For the thumb, groups A and C yield identical length ratios, while group B results in a longer distal phalanx. Group D produces equal lengths across all phalanges. For the fingers, all groups (A to D) produce different results. In groups A and B, the proximal phalanx is the longest, with group C producing the shortest distal phalanx. Group C yields a more balanced distribution compared to group A, and group D results in equal phalanx lengths.

\subsection{Analysis}

The optimization results indicate that assigning weighting coefficients does not necessarily lead to maximizing the corresponding evaluation factor. This is because the distribution of evaluation values within the feasible design space is not uniform. Additionally, each evaluation factor tends to bias the phalanx length ratios in different ways.

Workspace volume and overlap workspace volume are evaluated based on reachable points and voxel representation. The results show that a longer distal phalanx leads to a larger reachable region, resulting in a larger workspace volume. For overlap workspace, since it evaluates voxels simultaneously occupied by the thumb and other fingers, a longer distal phalanx is advantageous for generating a wider overlapping reachable region between opposing fingers.

Manipulability depends on joint motion ranges. When the motion ranges are identical, manipulability is more influenced by joints with larger motion ranges. However, this effect is limited to joints sharing the same motion direction. In the case of fingers, this corresponds to the three joints performing F/E motion. In the proposed model, the F/E motion ranges differ among the Metacarpophalangeal, Proximal Interphalangeal, and Distal Interphalangeal joints, which results in a tendency for manipulability to favor longer proximal phalanges. However, if identical motion ranges are assumed for all joints, the manipulability is maximized when all phalanx lengths are equal.

Sensitivity represents how much the fingertip position changes with respect to joint motion variations. Lower sensitivity indicates smaller positional changes for the same joint variation, enabling more precise control.

From the optimization results, it is observed that a shorter distal phalanx tends to reduce positional variation. In addition, within the proposed kinematic structure, large-scale motion is mainly contributed by the proximal and middle phalanges, whereas the distal phalanx plays a more significant role in fine manipulation. This observation is consistent with the relationship between link length and sensitivity in the proposed evaluation framework.

This result indicates that the finger structure inherently separates the roles of gross motion and fine manipulation, where proximal segments contribute to workspace expansion and distal segments contribute to precision control.

\section{Discussions}

Potential dexterity does not, by itself, guarantee the ability to grasp or manipulate objects. However, it can be useful for evaluating and referencing potential performance at the design stage without requiring information about specific objects or manipulation tasks. Furthermore, it can contribute to modifying kinematic parameters according to the design objective and prioritizing potential performance in specific directions.

In the proposed study, the evaluation was conducted based on the reachable workspace of the hand, thumb opposability through the overlap workspace, distal sensitivity, and global manipulability. During the optimization process, a minimum finger length was imposed, and a constraint was applied such that the phalanx lengths become shorter or equal toward the fingertip to resemble the human hand. In the process of extracting the maximum value for each factor, the minimum finger length was maintained while removing the phalanx length ordering constraints. Under this condition, the workspace volume was maximized when the distal phalanx was the longest. This indicates that a single-factor approach based solely on maximizing workspace volume is not appropriate.

Global manipulability was influenced by the joint range of motion. In cases where identical motions are generated—such as the three F/E motions of the fingers—when the joint ranges are equal, configurations with equal phalanx lengths were dominant. When the range of motion of a specific joint was larger, the phalanx closer to that joint tended to be longer. This indicates that the configuration of joint ranges of motion also affects the phalanx length ratios.

Finally, sensitivity represents the geometric responsiveness of the fingertip position to joint variations. A higher sensitivity corresponds to a larger displacement of the fingertip with respect to joint motion. When deriving the maximum value, the phalanx length ordering constraints were removed similarly to the workspace case, and the results showed that the distal phalanx became the longest. Therefore, when sensitivity is reduced, the distal phalanx becomes shorter, enabling finer motion at the fingertip joint, while the other phalanges become relatively longer, adjusting the structure toward configurations suitable for larger-scale motions.

In this study, not all weighting coefficients were exhaustively explored during the optimization process based on the evaluation factors. However, the results obtained by adjusting the weights were categorized into four groups. These optimization results indicate that, depending on which performance factor is prioritized, the resulting phalanx length ratios vary accordingly. This provides a useful reference for designers when determining the finger length ratios based on the intended performance.

Despite these findings, several limitations exist in this study. The present analysis is based purely on kinematic properties and does not consider contact mechanics, friction, or force transmission. Therefore, the proposed design implications should be interpreted as geometric and kinematic guidelines rather than complete predictors of grasp stability or force-based manipulation performance. In addition, the use of discretized joint sampling and voxel-based workspace approximation introduces resolution-dependent effects. While these may influence quantitative results, they are not expected to significantly affect the overall qualitative trends observed in this study. Furthermore, this study does not include geometric information such as finger thickness or width. For kinematic simplification in the analysis, the base positions of the fingers, except for the thumb, were assigned with the same height offset, and the lengths of the four fingers were set to be identical.

In future work, an extended model incorporating geometric factors such as finger thickness and width will be considered. In addition, evaluations will be conducted by allowing different finger lengths even under the optimized length ratios, and the effects of variations in the metacarpal joint positions and orientations on potential dexterity will be analyzed. Through this, it is expected that the proposed approach will contribute to constructing the kinematic structure of robotic hands during the design process and provide a more practical framework for kinematic parameter optimization in real robotic hand design.

\section{Conclusions}

This paper presented a framework for optimizing phalanx length ratios in a five-finger robotic hand based on potential dexterity. The proposed method evaluates candidate designs using global manipulability, workspace volume, overlap workspace volume, and sensitivity, and determines optimal configurations through a weighted objective formulation. The results demonstrate that the phalanx lengths do not contribute equally to the overall dexterity. Instead, the optimization reveals that each phalanx plays a distinct functional role within the kinematic structure. In particular, the distal segment significantly influences sensitivity and precision, while the proximal structure primarily contributes to workspace formation. In addition, the study shows that the selection of weighting coefficients does not directly guarantee the maximization of individual evaluation factors due to the non-uniform distribution of performance metrics within the feasible design space. Nevertheless, consistent structural tendencies observed across different weighting configurations provide meaningful design insights. The proposed framework offers a systematic approach to exploring design trade-offs between reachability, dexterity, and controllability. As a result, it can serve as a useful guideline for the kinematic design of multi-fingered robotic hands, particularly in the early stages of design where object-specific or force-based considerations are not yet incorporated.


\bibliographystyle{asmejour}   

\bibliography{bibtex} 

\end{document}